\def\year{2021}\relax
\documentclass[letterpaper]{article} 
\usepackage{comment}
\usepackage{pgfplots}
\pgfplotsset{width=10cm,compat=1.9}
\usepackage{subfig}
\usetikzlibrary{pgfplots.groupplots}
\usepackage[ruled,vlined,noend]{algorithm2e}
\usepackage{enumitem}
\usepackage{amsfonts}
\usepackage{amsmath}
\usepackage{booktabs,makecell}
\usepackage[export]{adjustbox}
\usepackage[switch]{lineno}

\DeclareMathOperator*{\argmax}{argmax} 
\DeclareMathOperator*{\argmin}{argmin} 

\newcommand*{\fnref}[1]{\textsuperscript{\ref{#1}}}

\usepackage{aaai21}  
\usepackage{times}  
\usepackage{helvet} 
\usepackage{courier}  
\usepackage[hyphens]{url}  
\usepackage{graphicx} 
\usepackage{natbib}  
\usepackage{caption} 
\title{Embracing Domain Differences in Fake News: Cross-domain Fake News Detection using Multimodal Data}
\author{
Amila Silva, 
Ling Luo,
Shanika Karunasekera,
Christopher Leckie
}
\affiliations{
    \\School of Computing and Information Systems\\ The University of Melbourne \\Parkville, Victoria, Australia\\
    \{amila.silva@student., ling.luo@, karus@, caleckie@\}unimelb.edu.au
}

\begin{document}

\maketitle

\begin{abstract}
With the rapid evolution of social media, fake news has become a significant social problem, which cannot be addressed in a timely manner using manual investigation. This has motivated numerous studies on automating fake news detection. Most studies explore supervised training models with different modalities (e.g., text, images, and propagation networks) of news records to identify fake news. However, the performance of such techniques generally drops if news records are coming from different domains (e.g., politics, entertainment), especially for domains that are unseen or rarely-seen during training. As motivation, we empirically show that news records from different domains have significantly different word usage and propagation patterns. Furthermore, due to the sheer volume of unlabelled news records, it is challenging to select news records for manual labelling so that the domain-coverage of the labelled dataset is maximized. Hence, this work: (1) proposes a novel framework that jointly preserves domain-specific and cross-domain knowledge in news records to detect fake news from different domains; and (2) introduces an unsupervised technique to select a set of unlabelled informative news records for manual labelling, which can be ultimately used to train a fake news detection model that performs well for many domains while minimizing the labelling cost. Our experiments show that the integration of the proposed fake news model and the selective annotation approach achieves state-of-the-art performance for cross-domain news datasets, while yielding notable improvements for rarely-appearing domains in news datasets.
\end{abstract}

\section{Introduction}
\textbf{Motivation.} Today, social media is considered as one of the leading and fastest media to seek news information online.
Thus, social media platforms provide an ideal environment to spread fake news (i.e., disinformation). Many times the cost and damage due to fake news are high and early detection to stop spreading such information is of importance. For example, it has been estimated that at least 800 people died and 5800 were admitted to hospital as a result of false information related to the COVID-19 pandemic, e.g., believing alcohol-based cleaning products are a cure for the virus\footnote{\url{https://www.bbc.com/news/world-53755067}}. Due to the high volumes of news generated on a daily basis, it is not practical to identify fake news using manual fact checking. Therefore, automatic detection of fake news has recently become a significant problem attracting immense research effort. 

\begin{figure}[t]
    \centering
    \subfloat[]{%
    \includegraphics[width=\linewidth]{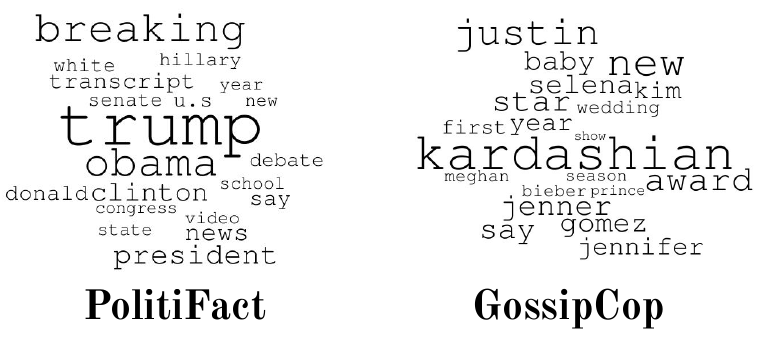}\label{fig:domain-specfic_text_results}%
    }
    \hspace{1cm}
    \subfloat[]{\adjustbox{width=\linewidth,valign=B,raise=0.1\baselineskip}{%
    \scriptsize
    \begin{tabular}{|c|c|c|c|c|}
    \hline
         Feature & \begin{tabular}[c]{@{}c@{}}Weiner \\ Index\end{tabular} & \begin{tabular}[c]{@{}c@{}}Network \\Depth\end{tabular}&\begin{tabular}[c]{@{}c@{}}Maximum\\Outdegree\end{tabular}&\begin{tabular}[c]{@{}c@{}} Propagation\\Speed\end{tabular}\\
         \hline
         p-value&1.81e-2 &5.81e-19&4.11e-4&3.42e-29\\
         \hline
    \end{tabular}
    \label{fig:domain-specfic_network_results}%
    }}
    \caption{(a) Word clouds for the top 20 words in PolitiFact and GossipCop. (b) Two-sample t-test results conducted using different graph-level features extracted from the propagation networks in PolitiFact and GossipCop.}\vspace{-2mm}
    \label{fig:domain-specfic_results}
\end{figure}

\textbf{Challenges.} Nevertheless, most existing fake news detection techniques fail to identify fake news in a real-world news stream for the following reasons. First, most existing techniques~\cite{silva-embedding-2020,zhou_safe_2020,shu_defend_2019,shu_hierarchical_2019,ruchansky_csi_2017} are trained and evaluated using datasets~\cite{shu_fakenewsnet_2018,cui2020coaid} that are limited to a single domain such as politics, entertainment, healthcare. However, a real news stream typically covers a wide variety of domains. We have empirically found that existing fake news detection techniques perform poorly for such a cross-domain news dataset despite yielding good results for domain-specific news datasets. This observation may be due to two reasons: (1) domain-specific word usage; and (2) domain-specific propagation patterns. For example, Figure~\ref{fig:domain-specfic_results} adopts two datasets from different domains, PolitiFact for politics and GossipCop for entertainment, which are two widely used labelled datasets to train fake news detection models. Fig.~\ref{fig:domain-specfic_results} shows that there are significant differences in the frequently used words and propagation patterns of these two datasets. To address this challenge, some previous works~\cite{wang_eann_2018,castelo2019topic} learned models to overlook such domain-specific information and only rely on cross-domain information (e.g., web-markup and readability features) for fake news detection. However, domain-specific knowledge could be useful for accurate identification of fake news. As a solution, this work aims to address \textbf{\textit{how to preserve domain-specific and cross-domain knowledge in news records to detect fake news in cross-domain news datasets.}} 
Second, the studies in~\cite{han2020graph,janicka2019cross} show that most fake news detection techniques are not good at identifying fake news records from unseen or rarely-seen domains during training. As a solution, fake news detection models can be learned using a dataset that covers as many domains as possible. Here we assume that the fake news detection model requires supervision as supervised techniques are known to be substantially better at identifying fake news compared to the unsupervised methods~\cite{yang2019unsupervised}. In such a supervised learning setting, each training (i.e., labelled) data point has an associated labelling cost. Thus, the total labelling budget constrains the number of data instances that can be selected for manual labelling. Due to the sheer volume of unlabelled news records available, there is a need to \textbf{\textit{identify informative news records to annotate such that the labelled dataset ultimately covers many domains while avoiding any selection biases.}} 

\textbf{Contribution.}
To address the aforementioned challenges, this work makes the following contributions:
\begin{itemize}[wide=0pt,noitemsep,topsep=0pt,parsep=0pt,partopsep=0pt]
\item We propose a multimodal\footnote{We define multimodality as information acquired from different sources/attributes following~\cite{zhang2017react}, instead of restricting just for sensory media (e.g., text, image).} fake news detection technique for cross-domain news datasets that learns domain-specific and cross-domain information of news records using two independent embedding spaces, which are subsequently used to identify fake news records. Our experiments show that the proposed framework outperforms state-of-the-art fake news detection models by as much as $7.55\%$ in F1-score.
\item We propose an unsupervised technique to select a given number of news records from a large data pool such that the selected dataset maximizes the domain coverage. By using such a dataset to train a fake news detection model, we show that the model achieves around $25\%$ F1-score improvements for rarely-appearing domains in news datasets.
\end{itemize}

\section{Related Work}

Fake news detection methods mainly rely on different attributes (text, image, social context) of news records to determine their veracity. Text content-based approaches~\cite{yang_hierarchical_2016,volkova_separating_2017,perez-rosas_automatic_2018,pennebaker_development_2015} mainly explore word usage and linguistic styles in the headline and body of news records to identify fake news. Some works analyse the images in news records along with the text content for fake news detection. For example, the studies in~\cite{jin2017multimodal,wang_eann_2018,khattar2019mvae} use pre-trained image models (e.g., VGG-19, ResNet) to extract features from images, which are integrated with text features to identify fake news. Also, some works consider the social context of a news record, i.e., how the record is propagated across social media, as another modality to differentiate fake news records from real ones. Existing work in this line mostly applies various machine learning techniques to extract features from propagation patterns, including Propagation Tree  Kernels~\cite{ma_detect_2017}, Recurrent Neural Networks~\cite{wu_tracing_2018,liu_early_2018}, and Graph Neural Networks~\cite{monti_fake_2019}. However, all these modalities (i.e., text, propagation patterns) generally show notable differences (see Figure~\ref{fig:domain-specfic_results}) for news records in different domains. Thus, most existing techniques perform poorly for cross-domain news datasets due to their inability to capture such domain-specific variations. Our model also relies on the text content and social context of news. However, 
the main objective of our model is to capture such domain-specific variations of news records.
\begin{figure*}
    \centering
    \includegraphics[width=\linewidth]{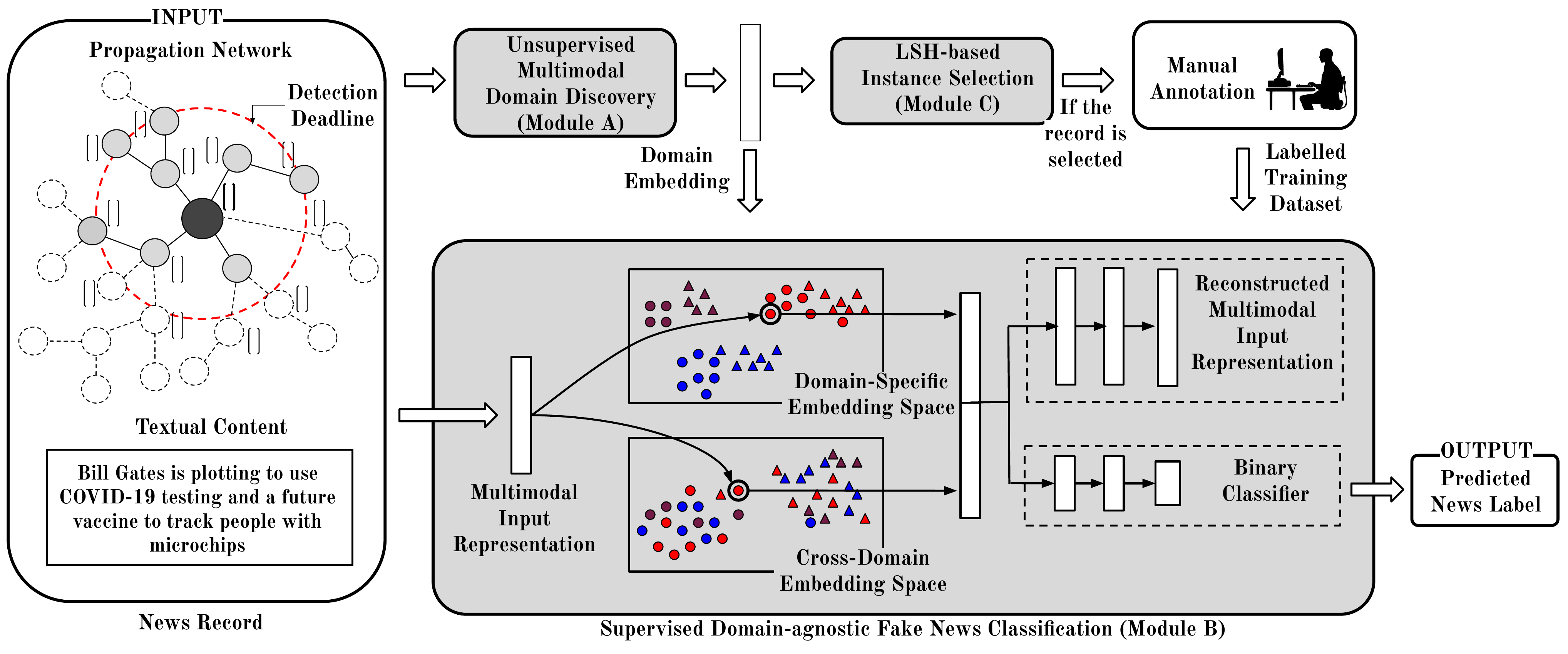}
    \caption{Overview of the proposed framework. In the illustrated embedding spaces, each data point's colour and shape denote its domain label and veracity label (i.e., triangle for fake news and circle otherwise) respectively.}
    \label{fig:overview}
    \vspace{-5mm}
\end{figure*}

\textbf{Domain-agnostic Fake News Detection.} 
Several previous works have attempted to perform fake news detection using cross-domain datasets. In~\cite{wang_eann_2018}, an event discriminator is learned along with a multimodal fake news detector to overlook domain-specific information in news records. The study in~\cite{castelo2019topic} carefully selects a set of features (e.g., psychological features, readability features) from news records that are domain-invariant. These techniques rely only on cross-domain information in news records. In contrast,~\citeauthor{han2020graph} (2020) consider cross-domain fake news detection as a continual learning task, which learns a model for a large number of tasks sequentially. This work adopts Graph Neural Networks to detect fake news using their propagation patterns and applies well-known continual learning approaches Elastic Weight Consolidation~\cite{kirkpatrick_overcoming_2017} and Gradient Episodic Memory~\cite{lopez-paz_gradient_2017} to address cross-domain fake news detection problem. This approach has two limitations: (1) it assumes that the news records from different domains arrive sequentially, though this is not always true for real-world streams; and (2) it requires the domain of news records to be known, which is not generally available. In contrast, our approach exploits both domain-specific and cross-domain knowledge of news records without knowing the actual domain of news records.

\textbf{Active Learning for Fake News Detection.}
Almost all the aforementioned models are supervised. Although there are unsupervised fake news detection techniques~\cite{yang_unsupervised_2019,hosseinimotlagh2018unsupervised}, they are generally inferior to the supervised approaches in terms of accuracy. However, the training of supervised models requires large labelled datasets, which are costly to collect. Therefore, how to obtain fresh and high-quality labelled samples for a given labelling budget is challenging. Some works~\cite{wang2020weak,bhattacharjee2017active} adopt conventional active learning frameworks to select high-quality samples, in which the model is initially trained using a small randomly selected dataset. Then, the beliefs derived from the initial model are used to select subsequent instances to annotate. This approach has two limitations: (1) it requires a pre-trained model to select instances; and (2) it is known to be highly vulnerable to the biases introduced by the initial model. In contrast, our instance selection approach does not depend on such an initial model. Also, none of the previous works attempted to explicitly maximize the domain-coverage of the labelled dataset, which is vital to train a model that perform equally well for multiple domains.

\begin{table}[]
    \centering
    \caption{Descriptive statistics of PolitiFact, GossipCop and CoAID datasets.}
    \begin{tabular}{c|c|c|c}
         Dataset &  PolitiFact & GossipCop & CoAID\\
         \hline
         \# Fake News & 269 & 1269 & 135\\
         \hline
         \# Real News & 230 & 2466 & 1568\\
         \hline
    \end{tabular}
    \label{tab:dataset_stat}
    \vspace{-5mm}
\end{table}

\section{Problem Statement}
Let $R$ be a set of news records. Each record $r \in R$ is represented as a tuple $\langle t^r, W^r, G^r\rangle$, where (1) $t^r$ is the timestamp when $r$ is published online; (2) $W^r$ is the text content of $r$; and (3) $G^r$ is the propagation network of $r$ for time bound $\Delta T$. We keep $\Delta T$ low (= five hours) for our experiments to evaluate early detection performance. 
Each propagation network $G^r$ is an attributed directed graph $(V^r, E^r, X^r)$, where nodes $V^r$ represent the tweets/retweets of $r$ and the edges $E^r$ represent the retweet relationships among them. $X^r$ is the set of attributes of the nodes (i.e., tweets) in $G^r$. More details about $E^r$ and $G^r$ are given in~\cite{supplement_aaai2021}.

Our problem consists of two sub-tasks: (1) select a set of instances $R^L$ from $R$ to label while adhering to the given labelling budget $B$, which constrains the number of instances in $R^L$. The labelling process assigns a binary label $y^r$ for each record $r$: $y^r$ is 1 if $r$ is false and 0 otherwise; (2) learn an effective model using $R^L$ to predict the label $y^r$ for unlabelled news records $r \in R^U$ as false or real news records. In this work, $R$ $(R^L \cup R^U)$ is not constrained to a specific domain. To emulate such a domain-agnostic dataset, we combine three publicly available datasets: (1) PolitiFact~\cite{shu_fakenewsnet_2018}, which consists of news related to politics; (2) GossipCop~\cite{shu_fakenewsnet_2018}, a set of news related to entertainment stories; and (3) CoAID~\cite{cui2020coaid}, a news collection related to COVID-19. All three datasets provide labelled news records and all the tweets related to each news item. The statistics of the datasets are shown in Table~\ref{tab:dataset_stat}.

\section{Our Approach}\label{sec:approach}

As shown in Fig.~\ref{fig:overview}, the proposed fake news detection model consists of two main components: (1) unsupervised domain embedding learning (Module A); and (2) supervised domain-agnostic news classification (Module B). These two components are integrated to identify fake news while exploiting domain-specific and cross-domain knowledge in the news records. In addition, the proposed instance selection approach (Module C) adopts the same domain embedding learning component to select informative news records for labelling, which eventually yields a labelled dataset that maximizes the domain-coverage.

\subsection{Unsupervised Domain Discovery}
For a given news record $r$, assume that its domain label is not available. The proposed unsupervised domain embedding learning technique exploits multimodal content (e.g., text, propagation network) of $r$ to represent the domain of $r$ as a low-dimensional vector $f_{domain}(r)$. Our approach is motivated by: (1) the tendency of users to form groups containing people with similar interests (i.e., homophily)~\cite{mcpherson2001birds}, which results in different domains having distinct user bases; and (2) the significant differences in domain-specific word usage as shown in Figure~\ref{fig:domain-specfic_text_results}.

We exploit the aforementioned motivations by constructing a heterogeneous network which consists of both users tweeting the news items and words in the news title as nodes, using the following steps (Line 1-9 in Algo.~\ref{algo:domain_emb_learning}): (1) create a set $S^r$ for each news record $r$ by adding all the users $U^r$ in the propagation network $G^r$ and all the words appearing in the news title $W^r$ (tokenized using whitespaces); (2) for each pair of items in $S^r$, build a weighted edge $e$ linking the two items in the graph; and (3)  repeat Steps 1 and 2 for all the news records, until we obtain the final network $G$. Then, we adopt the Louvain algorithm\footnote{Please see Supplementary Material in~\cite{supplement_aaai2021} for detailed pseudo code of the Louvain algorithm}~\cite{blondel2008fast} to identify communities in $G$. Here, we select the Louvain algorithm as it was shown to be one of the best performing parameter-free community detection algorithms in~\cite{fortunato2010community}. At the end of this step, we obtain a set of communities/clusters $C$, each having either a highly connected set of users or words. As the nodes of $G$ contain both users and words, such communities may have formed either due to a set of users engaging with similar news records or a set of words only appearing within a fraction of news records. Following the aforementioned motivations, this work assumes each community in $C$ belongs to a single domain. 

\begin{algorithm}[t]
 \LinesNumbered
 \SetAlgoLined
 \SetKwInput{Input}{Input}
 \SetKwInput{Output}{Output}
 \Input{A collection of news records $R$}
 \Output{Domain embeddings $f_{domain}(r)$ of $r \in R$}
 \tcp{Network construction}
 Initialize an empty graph $G$;\\
 \For{$r \in R$}{
 $S^r \leftarrow X^r \cup U^r$\\
    \For{each pair $(s_1, s_2) \in S$}{
        $e \leftarrow (\{s_1, s_2\}, 1)$;\\
        \uIf{edge $e$ exists in graph $G$}{
            Increment edge $e$ in graph $G$ by 1;
        }
        \uElse{
            Add edge $e$ to graph $G$;
        }
    }
}
\tcp{Community Detection}
$C \leftarrow$ Find communities in $G$ using Louvain;\\
\tcp{Embedding Learning} 
\For{$r \in R$}{
    Compute $f_{domain}(r)$ using Eq.~\ref{eq:domain_embedding}
}
Return $f_{domain}(r)$ of $r \in R$.
\caption{Domain Embedding Learning}\label{algo:domain_emb_learning}
\end{algorithm}

In the next step, we compute the soft membership $p(r \in c)$ of $r$ in a cluster $c$ using the following equation:
\begin{equation}
    p(r\in c) = \sum\limits_{v \in c \cap r} v_{deg}/\sum\limits_{c\in C}\sum\limits_{v \in r} v_{deg}
    \label{eq:likelihood_for_cluster_assingment}
\end{equation}

Here $p(r\in c)$ is proportional to the number of common users or words that $r$ and $c$ have. Each node (i.e., user or word) $v$ is weighted using the degree $v_{deg}$ in $G$ (i.e., number of occurrences) to reflect their varying importance for the corresponding community. Finally, we produce the domain embedding $f_{domain}(r) \in \mathbb{R}^{|C|}$ of $r$ as the concatenation of $r$'s likelihood belonging to communities in $C$:
\begin{equation}
    f_{domain}(r) = p(r\in c_1)\oplus p(r\in c_2)\oplus \dots p(r\in c_{|C|})
    \label{eq:domain_embedding}
\end{equation}
where $\oplus$ denotes concatenation.

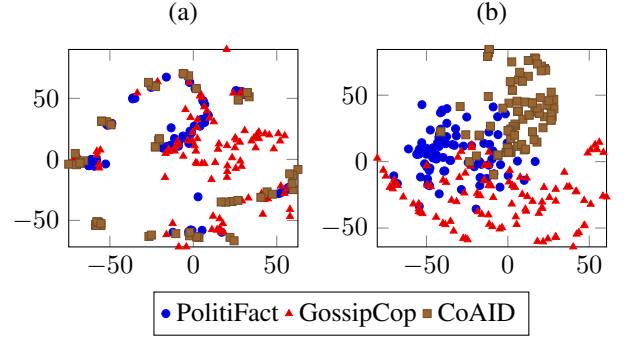
\begin{figure}[t]
\centering
 \begin{tikzpicture}
    \begin{groupplot}[group style = {group size = 2 by 2, horizontal sep = 30pt, vertical sep = 30pt}, width = 4.5cm, height = 4.3cm]
        \nextgroupplot[ title=(a), enlargelimits=false,
    width=0.55\linewidth,
    height=0.5\linewidth,
    legend style = {column sep = 0.5pt, legend columns = 3, legend to name = grouplegend,}]
\addplot+[
    only marks,
    mark=*,
    mark size=1.5pt]
table[]
{table2_1.txt};
\addplot+[
    only marks,
    mark=triangle*,
    mark size=1.5pt]
table[]
{table2_2.txt};
\addplot+[
    only marks,
    mark=square*,
    mark size=1.5pt]
table[]
{table2_3.txt};
    \legend{PolitiFact, GossipCop, CoAID}
        \nextgroupplot[ title=(b), enlargelimits=false,
    width=0.55\linewidth,
    height=0.5\linewidth]
\addplot+[
    only marks,
    mark=*,
    mark size=1.5pt]
table[]
{table3_1.txt};
\addplot+[
    only marks,
    mark=triangle*,
    mark size=1.5pt]
table[]
{table3_2.txt};
\addplot+[
    only marks,
    mark=square*,
    mark size=1.5pt]
table[]
{table3_3.txt};
     \end{groupplot}
    \node at ($(group c2r1) + (-2cm,-2.2cm)$) {\ref{grouplegend}};
\end{tikzpicture}
\caption{t-SNE visualization of domain embeddings from: 
(a) user-based domain discovery algorithm in~\cite{chen2020proactive} and (b) multimodal domain discovery approach proposed in this work.}
\label{fig:domain_embeddings}
\vspace{-5mm}
\end{figure}

In Figure~\ref{fig:domain_embeddings}, we adopt t-SNE~\cite{maaten2008visualizing} to visualize the domain embedding space of the proposed approach and the user-based domain discovery algorithm proposed in~\cite{chen2020proactive}. Due to space limitations, we present more details about the baseline in~\cite{supplement_aaai2021}. As can be seen in Figure~\ref{fig:domain_embeddings}, the proposed approach yields a clear separation between the domains compared to the baseline. This may be mainly due to the ability of our approach to jointly exploit multimodalities, both users and text of news records to discover their domains. In addition, most previous works on domain discovery ultimately assign hard domain labels for news records, which could lead to substantial information loss. For example, some news records may belong to multiple domains, which cannot be captured using hard domain labels.  Hence, by having a low-dimensional vector to represent embedding, our approach could preserve such knowledge related to the domains of news records.

\subsection{Domain-agnostic News Classification}

In our news classification model, each news record $r$ is represented as a vector $f_{input}(r)$ using the textual content $W^r$ and the propagation network $G^r$ of $r$ (elaborated in the section Experiments). Then, our classification model maps $f_{input}(r)$ into two different subspaces such that one preserves the domain-specific knowledge, $f_{specific}: f_{input}(r)\rightarrow \mathbb{R}^d$, and the other preserves the cross-domain knowledge $f_{shared}: f_{input}(r)\rightarrow \mathbb{R}^d$, of $r$. Here $d$ is the dimension of the subspaces. Then, the concatenation $f_{specific}(r)$ and $f_{shared}(r)$ is used to recover the label $y^r$ and the input representation $f_{input}(r)$ of $r$ during training via two decoder functions $g_{pred}$ and $g_{recons}$ respectively.
\begin{align}
    \overline{y^r} &= g_{pred}(f_{specific}(r)\oplus f_{shared}(r))\notag\\
    \overline{f_{input}(r)} &= g_{recon}(f_{specific}(r) \oplus f_{shared}(r))\notag\\
    L_{pred} &= BCE(y^r, \overline{y^r})\\
    L_{recon} &= ||f_{input}(r)-\overline{f_{input}(r)}||^2
\end{align}
where $\overline{y^r}$ and $\overline{f_{input}(r)}$ denote the predicted label and the predicted input representation respectively. $BCE$ stands for the Binary Cross-Entropy loss function. We minimize $L_{pred}$ and $L_{recon}$ to find the optimal parameters of $(f_{specific}, f_{shared}, g_{pred}, g_{recon})$. 

However, $L_{pred}$ and $L_{recon}$ do not leverage domain differences in news records. Hence, we now discuss how the mapping functions for subspaces, $f_{specific}$ and  $f_{shared}$, are further learned to preserve the domain-specific and cross-domain knowledge in news records. 



\subsubsection{Leveraging Domain-specific Knowledge}
To preserve the domain-specific knowledge, we introduce an auxiliary loss term $L_{specific}$ to learn a new decoder function $g_{specific}$ to recover the domain embedding $f_{domain}(r)$ of $r$ using the domain-specific representation $f_{specific}(r)$. We minimize $L_{specific}$ to find the optimal parameters for $(f_{specific}, g_{specific})$ to capture the domain-specific knowledge by $f_{specific}$, and this process can be defined as follows: 
\begin{align}
    L_{specific} = ||f_{domain}(r)-g_{specific}(f_{specific}(r))||^2\notag\\
    (\hat{g}_{specific}, \hat{f}_{specific}) = \argmin_{(g_{specific}, f_{specific})}(L_{specific})\label{eq:domain-specific}
\end{align}

\subsubsection{Leveraging Cross-domain Knowledge}
In contrast, we learn $f_{shared}$ to overlook domain-specific knowledge of the news records. Consequently, $f_{shared}$ preserves the cross-domain knowledge in the news records. Here, we train a decoder function $g_{shared}$ to accurately predict the domain of $r$ using $f_{shared}(r)$. Meanwhile, we learn $f_{shared}$ to fool the decoder $g_{shared}$ by maximizing the loss of $g_{shared}$. Such a formulation forces $f_{shared}$ to only rely on cross-domain knowledge, which are useful to transfer the knowledge across domains. This process can be defined as a minimax game between $g_{shared}$ and $f_{shared}$ as follows: 
\begin{align}
    L_{shared} = ||g_{shared}(f_{shared}(r)) - f_{domain}(r)||^2\notag\\
    (\hat{g}_{shared}, \hat{f}_{shared}) = \argmin_{f_{shared}}\argmax_{g_{shared}} (-L_{shared})\label{eq:domain-shared}
\end{align}
\subsubsection{Integrated Model}
Then the final loss function of the model is formulated as:
\begin{equation}
    L_{final} = L_{pred} + \lambda_1L_{recon} + \lambda_2L_{specific}- \lambda_3L_{shared}
\end{equation}
where $\lambda_1, \lambda_2$ and $\lambda_3$ controls the importance given to each loss term compared to $L_{pred}$ (i.e., main task). 

To learn the minimax game in $L_{shared}$, the final loss function $L_{final}$ is sequentially optimized using the following two steps: 
\begin{align}
    (\widehat{\theta_1}) &= \argmin_{\theta_1} L_{final}(\theta_1, \theta_2)\\
    (\widehat{\theta_2}) &= \argmax_{\theta_2} L_{final}(\widehat{\theta_1}, \theta_2)
\end{align}
where $\theta_1$ and $\theta_2$ denote the parameters in $(f_{specific}$, $f_{shared}$, $g_{specific}$, $g_{pred}$, $g_{recon})$ and $g_{shared}$ respectively. The empirically studied convergence properties of the proposed optimization scheme are presented in~\cite{supplement_aaai2021}. 



\subsection{LSH-based Instance Selection}
The aforementioned model is able to exploit the domain-specific and cross-domain knowledge in news records to identify their veracity. Nevertheless, if the model is used to identify fake news records in unseen or rarely appearing domains during training, we empirically observe that the performance of the model substantially drops. This observation is expected and is consistent with the findings in~\cite{castelo2019topic}, which could be due to the domain-specific word usage and propagation patterns as shown in Fig.~\ref{fig:domain-specfic_results}. Hence, we propose an unsupervised technique to come up with a labelled training dataset for a given labelling budget $B$ such that it covers as many domains as possible. The ultimate objective of this technique is to learn a model using such a dataset that performs well for many domains.

Our approach initially represents each news record $r\in R$ using its domain embedding $f_{domain}(r)$. Then, we propose a Locality-Sensitive Hashing (LSH) algorithm based on random projection to select a set of records in $R$ that are distant in the domain embedding space, which can be elaborated using the following steps:

 \begin{enumerate}[wide=0pt,noitemsep,topsep=0pt,parsep=0pt,partopsep=0pt]
    \item Create $|H|$ different hash functions such as $H_i(r) = sgn(h_i\cdot f_{domain}(r))$, where $i\in\{0,1, \ldots,|H|-1\}$ and $h_i$ is a random vector, and $sgn(.)$ is the sign function. The random vectors $h_i$ are generated using the following probability distribution, as such a distribution was shown to perform well for random projection-based techniques~\cite{achlioptas2001database}:
    \begin{equation}
    h_{i,j} = \sqrt{3}\times \begin{cases}
    +1  & \text{with probability $1/6$} \\
    0   & \text{with probability $2/3$} \\
    -1  & \text{with probability $1/6$}
    \end{cases}
    \label{eq:random_distribution}
    \end{equation}
    \item Construct an $|H|$-dimensional hash value for each news record $r$ as $H_0(r) \oplus H_1(r) \oplus \ldots \oplus H_{|H|-1}(r)$, where $\oplus$ defines the concatenation operation. According to the Johnson-Lindenstrauss lemma~\cite{johnson1984extensions}, such hash values approximately preserve the distances between the news records in the original embedding space with high probability. Hence, neighbouring records in the domain embedding space are mapped to similar hash values.
    \item Group the news records with similar hash values to construct a hash table. 
    \item Randomly pick a record from each bin in the hash table and add to the selected dataset pool.
    \item Repeat steps (1), (2), (3) and (4) until the size of the dataset pool reaches the labelling budget $B$.
\end{enumerate}

\makeatletter
\newcommand\resetstackedplots{
\makeatletter
\pgfplots@stacked@isfirstplottrue
\makeatother
\addplot [forget plot,draw=none] coordinates{(PLT,0) (GSP,0) (CVD,0)};
}

\definecolor{blizzardblue}{rgb}{0.67, 0.9, 1}
\definecolor{coral}{rgb}{1.0, 0.7, 0.31}
\definecolor{darkred}{rgb}{0.55, 0.0, 0.0}

\begin{figure}[t]
\centering
\hspace{-0.5cm}
\subfloat[]{%
 \begin{tikzpicture}
    \begin{axis}[
    ybar stacked,
    enlargelimits=0.15,
    legend style={at={(0.5,-0.20)},
    anchor=north,legend columns=2},
    ylabel={\# Instances},
    symbolic x coords={PLT, GSP, CVD},
    xtick=data,
    width = 0.5\linewidth,
    height = 4cm,
    bar width =6pt,
    ytick={100,200,300,400},ymax=400]

    \addplot+[ybar, color=blue, bar shift=-4pt, bar width = 8pt] plot coordinates {(PLT,22) (GSP,91) (CVD,12)};
    \addplot+[ybar, color=blue!25, bar shift=-4pt, bar width = 8pt] plot coordinates {(PLT,19) (GSP,187) (CVD,114)};
    \resetstackedplots
    \addplot+[color=red, bar shift=4pt, bar width = 8pt] plot coordinates {(PLT,52) (GSP,101) (CVD,39)};
    \addplot+[color=red!25, bar shift=4pt, bar width = 8pt] plot coordinates {(PLT,61) (GSP,119) (CVD,82)};
    \legend{Rand-Fake,Rand-Real,LSH-Fake, LSH-Real}
    \end{axis}
\end{tikzpicture}%
\label{fig:stat_count}
}
\hspace{-0.5cm}
\subfloat[]{%
 \begin{tikzpicture}
    \begin{axis}[
		xlabel=$B/|R|$,
		ylabel=$\lambda$,
		width=0.55\linewidth,
		xtick={0.1, 0.3, 0.5, 0.7, 0.9},
		legend style={at={(0.5,-0.40)},
    anchor=north,legend columns=2}]
	\addplot[color=blue,mark=x] coordinates {
		(0.1,1.87)
		(0.2,1.92)
		(0.3,1.97)
		(0.4,2.07)
		(0.5,2.12)
		(0.6,2.14)
		(0.7,2.15)
		(0.8,2.17)
		(0.9,2.27)
	};
	\addplot[color=red,mark=*] coordinates {
		(0.1,1.14)
		(0.2,1.32)
		(0.3,1.43)
		(0.4,1.54)
		(0.5,1.67)
        (0.6,1.74)
        (0.7,1.87)
        (0.8,2.00)
        (0.9,2.13)
	};
	\legend{Rand,LSH}
	\end{axis}
\end{tikzpicture}
\label{fig:stat_lambda}
}

\caption{Statistics of datasets selected using random selection (Rand) and the proposed LSH-based technique (LSH). (a) Number of fake and real news records selected from each domain when $B/|R|=0.1$ and (b) domain-coverage measure $\lambda$ (lower $\lambda$ is better) for different $B/|R|$ values.}
\label{fig:selected_instances}
\vspace{-5mm}
\end{figure}
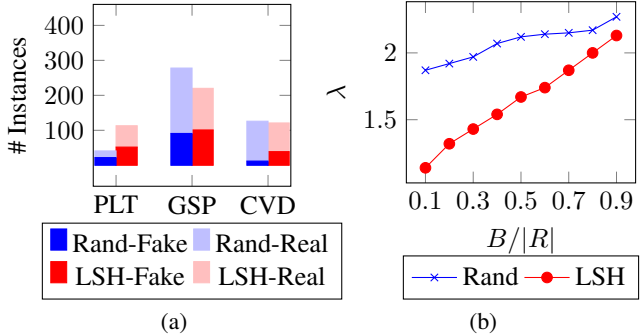

In Figure~\ref{fig:stat_count}, we compare $10\%$ of the original dataset selected using the proposed approach and random selection. As can be seen, random selection follows the empirical distribution of the datasets in Table~\ref{tab:dataset_stat} and picks few instances from rarely appearing domains (e.g., fake/real news in PolitiFact, fake news in CoAID). Thus, the model trained on such a dataset may poorly perform on rarely appearing domains. In contrast, the proposed approach provides a significant number of samples from even rarely occurring domains.

In addition, the proposed approach is efficient ($O(|H||R|)$ complexity) compared to the naive farthest point selection algorithms (e.g., k-Means~\cite{lloyd1982least} with $O(|R|^2)$ complexity, where $|R| >> |H|$). To measure the domain coverage of the instances selected from the proposed instance selection approach, we adopt the metric introduced in~\cite{laib2017unsupervised}, which can be computed as follows for a given set of records ${r_1, r_2, ..., r_n}$ that are represented using their domain embeddings: $\lambda = \frac{1}{\overline{\delta}}(\frac{1}{n}\sum_{i=1}^n(\delta_i-\overline{\delta})^2)^{\frac{1}{2}}$,
where $\delta_i=min_k (L2\text{ }norm(f_{domain}(r_i), f_{domain}(r_k)))$ and $\overline{\delta} = \sum\limits \delta_i/n$. If the coverage is high, $\lambda$ is small. Hence, the proposed approach yields a better domain-coverage compared to random instance selection as shown in Figure~\ref{fig:stat_lambda}. 



\section{Experiments}
\subsection{Experimental Setup}

\subsubsection{Encoding and Decoding Functions}
In our model, each record $r$ is initially represented as a low-dimensional vector $f_{input}(r)$ using its text content and propagation network. We adopt RoBERTa-base, a robustly optimized BERT pre-training model~\cite{liu2019roberta} to learn the text-based representation $f_{text}(r)$ of $r$. The propagation network-based representation $f_{network}(r)$ of $r$ is represented using the unsupervised network representation learning technique proposed in~\cite{silva-embedding-2020}. Then, the final input representation $f_{input}(r)$ is constructed as $f_{text}(r) \oplus f_{network}(r)$, where $\oplus$ denotes concatenation. All the other encoding and decoding functions, $(f_{specific}$, $f_{shared}$, $g_{specific}$, $g_{shared}$, $g_{pred}$, $g_{recon})$, are modelled as 2-layer feed-forward networks with sigmoid activation\footnote{\label{note1}We present more details about implementations and parameter selections in the Supplementary Material in~\cite{supplement_aaai2021}}. 

\subsubsection{Dataset} We combine three disinformation datasets: (1) PolitiFact; (2) GossipCop; and (3) CoAID, to produce a cross-domain news dataset\footnote{Here we do not consider the existing datasets on rumour detection~\cite{kochkina-etal-2018-one,ma_detect_2017} as they are not consistent with the fake news definition (i.e., disinformation).}. Then, we randomly choose 75\% of the dataset as the candidate data pool $R_{pool}$ for training and the remaining 25\% for testing. For a given labelling budget $B$, we select $B$ instances from $R_{pool}$ to train the model. The same process is performed for 3 different training and test splits and the average performance is reported. We evaluate the performance for each domain separately using the testing instances from each domain. For the evaluation, we adopt four metrics: (1) Accuracy (Acc); (2) Precision (Prec); (3) Recall (Rec); and (4) F1 Score (F1).

\subsubsection{Baselines} In Table~\ref{tab:results1}, we compare our approach with seven widely used fake detection techniques and their variants\fnref{note1}.

\subsubsection{Parameter Settings}After performing a grid search, we have set the hyper-parameters in our model as\fnref{note1}: $\lambda_1=1$, $\lambda_2=10$, $\lambda_3=5$, $d=512$. 
To satisfy the Johnson–Lindenstrauss lemma, we set $|H|=10\text{ }(>> log(|R|)$. For the specific parameters of the baselines, we use the default parameters mentioned in their original papers.

\subsection{Results}

\begin{table*}[t]
    \scriptsize
    \centering
    \caption{Results for fake news detection of different methods, which are classified under three categories: (1) text content-based approaches (T); (2) social context-based approaches (S); and (3) multimodal approaches (M).}
    \begin{tabular}{|p{3.8cm}|c|c|c|c|c|c|c|c|c|c|c|c|c|c|c|}
        \hline
         Method & \multicolumn{3}{c|}{Type}& \multicolumn{4}{c|}{Politifact} &\multicolumn{4}{c|}{Gossipcop}&\multicolumn{4}{c|}{CoAID}\\ 
         \cline{2-16}
         &T&S&M& Acc & Prec & Rec & F1& Acc & Prec & Rec & F1 &Acc & Prec & Rec & F1\\
         \hline
         LIWC~\cite{pennebaker_development_2015} &\checkmark&&&0.488&0.680&0.565&0.432&0.662&0.550&0.516&0.472&0.903&0.586&0.531&0.538\\
         text-CNN~\cite{kim_convolutional_2014} &\checkmark&&&0.608&0.621&0.623&0.608&0.733&0.698&0.703&0.701&0.903&0.679&0.674&0.677\\
         HAN~\cite{yang_hierarchical_2016} &\checkmark&&&0.632&0.672&0.651&0.648&0.716&0.703&0.709&0.706&0.919&0.698&0.682&0.688\\
         EANN-Unimodal~\cite{wang_eann_2018} &\checkmark&&&0.794&0.811&0.790&0.791&0.765&0.732&0.738&0.734&0.925&0.842&0.763&0.792\\
         \hline
         HPNF~\cite{shu_hierarchical_2019} &&\checkmark&&0.697&0.692&0.683&0.687&0.721&0.703&0.689&0.695&0.902&0.652&0.693&0.672\\
         AE~\cite{silva-embedding-2020} &&\checkmark&&0.784&0.783&0.774&0.779&0.834&0.828&0.802&0.812&0.928&0.686&0.673&0.677\\
         \hline
         HPNF + LIWC~\cite{shu_hierarchical_2019} &&&\checkmark&0.704&0.723&0.708&0.716&0.734&0.715&0.706&0.708&0.911&0.682&0.709&0.690\\
         SAFE~\cite{zhou_safe_2020}&&&\checkmark&0.793&0.782&0.771&0.775&0.831&0.822&0.798&0.806&0.931&0.754&0.744&0.748\\
         EANN-Multimodal~\cite{wang_eann_2018}&&&\checkmark&0.804&0.808&0.794&0.798&0.836&0.812&0.815&0.813&0.944&0.849&0.803&0.808\\
         \hline
         \hline
         Our Approach ($B=100\%|R_{pool}|$)&&&\checkmark&\bf 0.840&\bf0.836&\bf0.831&\bf0.835&\bf0.877&\bf0.840&\bf0.832&\bf0.836&\bf0.970&\bf0.876&\bf0.863&\bf0.869\\
         Our Approach ($B=50\%|R_{pool}|$)&&&\checkmark&\bf 0.838 &\bf 0.836 &\bf 0.828&\bf0.833&\bf0.848&\bf0.822&\bf0.797&\bf 0.808 &\bf0.963&\bf0.870&\bf 0.854&\bf 0.862\\
         \hline
         \multicolumn{4}{|l|}{\textbf{Ablation Study ($B=100\%|R_{pool}|$)}} &&&&&&&&&&&&\\
         \multicolumn{4}{|l|}{{(-) \textit{Domain-shared loss}}}&0.823&0.821&0.812&0.815&0.864&0.832&0.828&0.829&0.956&0.857&0.861&0.858\\
         \multicolumn{4}{|l|}{{(-) \textit{Domain-specific loss}}}&0.792&0.800&0.783&0.786&0.858&0.832&0.821&0.828&0.934&0.850&0.857&0.853\\
         \multicolumn{4}{|l|}{{(-) \textit{Network modality}}}&0.816&0.815&0.817&0.815&0.765&0.749&0.745&0.746&0.945&0.803&0.855&0.827\\
         \multicolumn{4}{|l|}{{(-) \textit{Text modality}}}&0.804&0.798&0.793&0.795&0.837&0.835&0.815&0.817&0.932&0.711&0.704&0.707\\
         \hline
    \end{tabular}
    \label{tab:results1}
    \vspace{-4mm}
\end{table*}

\subsubsection{Quantitative Results for Fake News Detection}
As shown in Table~\ref{tab:results1}, the proposed approach yields substantially better results for all three domains, outperforming the best baseline by as much as $7.55\%$ in F1-score. The best baseline, EANN-Multimodal, also adopts domain-information when determining fake news. This observation shows the importance of having domain-knowledge of news records when identifying fake news in cross-domain datasets. In addition to the architectural differences of the model, EANN-Multimodal is different from our approach for two reasons: (1) EANN-Multimodal only preserves cross-domain knowledge in news records. Thus, it overlooks domain-specific knowledge, which is shown to be useful in our ablation study in Table~\ref{tab:results1}; and (2) EANN-Multimodal adopts a hard label (i.e., exclusive membership) to represent the domain of a news record. Our approach conversely uses a vector to represent the domain of a news record. Thus, our approach can accurately represent the likelihood of each record for different domains. These differences may explain the importance of our approach compared to the best baseline.

Out of the baselines, the multimodal approaches (except HPNF+LIWC) generally achieve better results compared to the uni-modal approaches. Thus, we can conclude that each modality (i.e., propagation network and text) of news records provides unique knowledge for fake news detection. In HPNF+LIWC, each news record is represented using a set of hand-crafted features. In contrast, other multimodal approaches including our approach learn data-driven latent representations for news records, which may be able to capture latent and complex information in news records that are useful to determine fake news. These observations further support two main design decisions in our model: (1) to exploit multimodalities of news records; and (2) to adopt a representation learning-based technique.

\subsubsection{Ablation Study}
Our ablation study in Table~\ref{tab:results1} shows that without the domain-specific loss (Eq.~\ref{eq:domain-specific}) and the cross-domain loss (Eq.~\ref{eq:domain-shared}), the F1-score of the model substantially drops by around $6\%$ and $3\%$ for the PolitiFact dataset, which is the smallest domain of the training dataset. Hence, it is important to have a domain-specific layer to preserve the domain-specific knowledge and a separate cross-domain layer to transfer common knowledge between domains.

\begin{figure}[t]
\centering
 \begin{tikzpicture}
    \begin{groupplot}[group style = {group size = 3 by 1, horizontal sep = 30pt}, width = 4.5cm, height = 4.3cm]
        \nextgroupplot[ enlargelimits=false,
    width=0.55\linewidth,
    height=0.5\linewidth,
    legend style = {column sep = 0.5pt, legend columns = 3, legend to name = grouplegend,},
    title=(a)]
\addplot+[
    only marks,
    mark=*,
    mark size=1.5pt]
table[]
{table5_1.txt};
\addplot+[
    only marks,
    mark=triangle*,
    mark size=1.5pt]
table[]
{table5_2.txt};
\addplot+[
    only marks,
    mark=square*,
    mark size=1.5pt]
table[]
{table5_3.txt};
    \legend{PolitiFact, GossipCop, CoAID}
        \nextgroupplot[ enlargelimits=false,
    width=0.55\linewidth,
    height=0.5\linewidth,
    title=(b)]
\addplot+[
    only marks,
    mark=*,
    mark size=1.5pt]
table[]
{table4_1.txt};
\addplot+[
    only marks,
    mark=triangle*,
    mark size=1.5pt]
table[]
{table4_2.txt};
\addplot+[
    only marks,
    mark=square*,
    mark size=1.5pt]
table[]
{table4_3.txt};
    \end{groupplot}
    \node at ($(group c2r1) + (-2.5cm,-2.1cm)$) {\ref{grouplegend}};
\end{tikzpicture}
\caption{t-SNE visualization of the (a) domain-specific and (b) cross-domain embeddding spaces.}
\label{fig:latent_spaces}
\vspace{-5mm}
\end{figure}
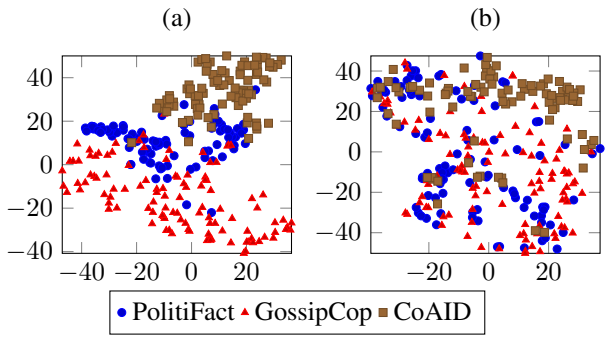

To check whether our model actually learns the aforementioned intuition behind each embedding layer, we visualize each embedding layer using t-SNE in Figure~\ref{fig:latent_spaces}. As can be seen, the domain-specific embedding layer preserves the domain of the news records by mapping different domains into different clusters. In contrast, we cannot identify the domain labels of news records from the cross-domain embedding space. Hence, this embedding space is useful to share common knowledge between records from different domains. 

Furthermore, we analyse the contribution of each modality. It can be seen that \textit{network modality} is more useful to determine fake news in GossipCop, while 
\textit{text modality} is the most informative one for CoAID. This observation further signifies the importance of multimodal approaches to train models that generalize for multiple domains.

\subsubsection{Evaluation of LSH-based Instance Selection}
As shown in Table~\ref{tab:results1}, our model outperforms the baselines even with a constrained budget $B$ ($50\% |R_{pool}|$) to select training data using the LSH-based instance selection technique. To verify its significance further, Figure~\ref{tab:different_labelling_budget} compares the proposed LSH-based instance selection approach with random instance selection for different $B$ values. The proposed approach substantially outperforms the random instance selection for the rarely-appearing or highly imbalanced domains. It increases F1-score by $24\%$ for PolitiFact and $27\%$ for CoAID, when $B/|R_{pool}| = 0.1$. This may be due to the ability of our approach to maximize the coverage of domains when selecting instances (see Figure~\ref{fig:selected_instances}), instead of biasing towards a domain with larger number of records.

\definecolor{darkbrown}{rgb}{0.4, 0.26, 0.13}
\begin{figure}[t]
\centering
 \begin{tikzpicture}
    \begin{groupplot}[group style = {group size = 3 by 1, horizontal sep = 40pt}, width = 4.5cm, height = 4.1cm]
        \nextgroupplot[ title = Random Selection,
            legend style = {column sep = 0.5pt, legend columns = 3, legend to name = grouplegend,}, ytick={0.0, 0.2, 0.4, 0.6,0.7, 0.7, 0.8, 0.9, 1.0},xtick= {0.1, 0.2, 0.3, 0.4, 0.5}, xlabel=$B/|R_{pool}|$,ymax=0.92, ymin=0.52, ylabel=F1-score]
\addplot[
    color=blue,
    mark=*,
    ]
    coordinates {
    (0.1,0.58)(0.2,0.67)(0.3,0.70)(0.4,0.74)(0.5,0.77)
    };
\addplot[
    color=red,
    mark=triangle*,
    ]
    coordinates {
    (0.1,0.78)(0.2,0.80)(0.3,0.83)(0.4,0.82)(0.5,0.84)
    };
\addplot[
    color=darkbrown,
    mark=square*,
    ]
    coordinates {
    (0.1,0.63)(0.2,0.74)(0.3,0.77)(0.4,0.81)(0.5,0.85)
    };
    \legend{Politifact, GossipCop, CoAID}
        \nextgroupplot[
        title= LSH-based Selection,
        ytick={0.0, 0.2, 0.4, 0.6, 0.7, 0.8, 0.9, 1.0},
        xtick = {0.1, 0.2, 0.3, 0.4, 0.5},
        ymax=0.92, ymin=0.52,
        xlabel=$B/|R_{pool}|$, ylabel=F1-score]
\addplot[
    color=blue,
    mark=*,
    ]
    coordinates {
    (0.1,0.72)(0.2,0.74)(0.3,0.78)(0.4,0.80)(0.5,0.83)
    };
\addplot[
    color=red,
    mark=triangle*,
    ]
    coordinates {
    (0.1,0.77)(0.2,0.79)(0.3,0.80)(0.4,0.83)(0.5,0.85)
    };
\addplot[
    color=darkbrown,
    mark=square*,
    ]
    coordinates {
    (0.1,0.80)(0.2,0.82)(0.3,0.82)(0.4,0.84)(0.5,0.87)
    };
    \end{groupplot}
    \node at ($(group c2r1) + (-2.7cm,-2.6cm)$) {\ref{grouplegend}};
\end{tikzpicture}
\caption{F1-scores for the fake news detection task with different instance selection strategies.}
\label{tab:different_labelling_budget}
\vspace{-5mm}
\end{figure}
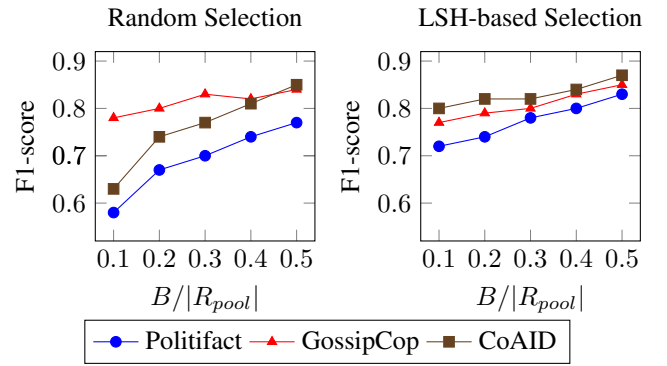


\section{Conclusion}
In this work, we proposed a novel fake news detection framework, which exploits domain-specific and cross-domain knowledge in news records to determine fake news from different domains. Also, we introduced a novel unsupervised approach to select informative instances for manual labelling from a large pool of unlabelled news records. The selected data pool is subsequently used to train a model that can perform equally for different domains. The integration of the aforementioned two contributions yields a model with low labelling budgets that outperforms existing fake news detection techniques by as much as $7.55\%$ in F1-score.

For future work, we intend to extend our model as an online learning framework to determine fake news in a real-world news stream, which typically covers a large number of domains. This setting introduces new challenges such as capturing newly emerging domains and handling temporal changes in domains. Also, how to use the alignment in multimodal information to weakly guide the learning process of the proposed model is another interesting direction to explore, which may further reduce the labelling cost in a conventional supervised learning setting. 

\section*{Acknowledgments}
This research was financially supported by Melbourne Graduate Research Scholarship and Rowden White Scholarship. We would like to specially thank Yi Han for his insightful comments and suggestions for this work. We are also grateful for the time and effort of the reviewers in providing valuable feedback on our manuscript. 

\bibliography{AAAI_draft.bib}

\end{document}


\maketitle

\begin{abstract}
This is the supplementary material for the paper titled "Embracing Domain Differences in Fake News: Cross-domain Fake News Detection using Multimodal Data". 
\end{abstract}

\section{Louvain Algorithm for Community Detection}
This section presents more details about the Louvain algorithm~\cite{blondel2008fast}, which is used in the proposed domain embedding learning approach to identify communities in a network. 

As shown in Algorithm~\ref{algo:louvain_algorithm}, the Louvain algorithm identifies the communities in a network using the following steps:
\begin{enumerate}
    \item Each vertex is placed in their own community (Line 1 in Algo.~\ref{algo:louvain_algorithm});
    \item Each vertex is retained in its own cluster or merge with an immediate neighbour such that the modularity scores of the network is maximised (Line 3-15 in Algo.~\ref{algo:louvain_algorithm}). The modularity score is computed as:
    \begin{align*}
        Q = \bigg[\frac{\sum_{in} +k_{i,in}}{2m} - \bigg(\frac{\sum_{tot} +k_{i}}{2m}\bigg)^2\bigg] -\\ \bigg[\frac{\sum_{in}}{2m} - \bigg(\frac{\sum_{tot}}{2m}\bigg)^2 - \bigg(\frac{k_i}{2m}\bigg)^2\bigg]
    \label{eq:mod}
    \end{align*}
    where $\sum_{in}$ and $\sum_{tot}$ represents the total weight of all links inside a community/cluster and total weight of all links to a community/cluster, respectively. Similarly, the terms $k_i$ and $k_{i,in}$ denote the total weight of all links to $i$ and total weight of links to $i$ within the community/cluster. Lastly, $m$ denotes the total weight of all links in the network graph;
    \item Build a new network where vertices in the same community are combined as a single vertex.
     \item Repeat Steps 2 and 3 until there are no more mergings between communities.   
\end{enumerate}

At the end of this algorithm, we will obtain a set of communities of the provided network such that the modularity score of the network is maximised. We selected this algorithm in our model because it is known~\cite{lim2017clustop} to generate a relatively small number of communities compared to other parameter-free community detection algorithms such as Infomap~\cite{rosvall2008maps} and Label Propagation~\cite{raghavan2007near}.

\begin{algorithm}[t]
 \LinesNumbered
 \SetAlgoLined
 \SetKwInput{Input}{Input}
 \SetKwInput{Output}{Output}
 \SetKwRepeat{Do}{do}{while}
 \Input{$G = (V, E)$ where $V$ and $E$ are the vertices and edges of the network $G$}
 \Output{$A = (V, C)$: Assignment of vertices $V$ into communities $C$}
 Assign all vertices $v$ into their own community;\\
 \Do{$A$ stabilises (i.e., no more shifts)}{
 \For{$v \in V$}{
 $MaxModularity \leftarrow -1$;\\
 $MaxModNeighbour \leftarrow NULL$;\\
 \For{each neighbour $v_n$ of $v$}{
 $ShiftMod \leftarrow$ Modularity score of shifting $v$ to $v_n$’s community;\\
 \If{$ShiftMod > MaxModularity$}{
 $MaxModularity \leftarrow ShiftMod;$\\
 $MaxModNeighbour \leftarrow v_n$;\\ 
 }
 }
 $OriginalMod \leftarrow $Modularity score of $v$ in its original community;\\
 \If{$OriginalMod > MaxModularity$}{
    Shift $v$ to the community of $MaxModNeighbour$;\\
 }
 \Else{
    Keep $v$ in its original community;\\
 }
 }
 }
\caption{Louvain Algorithm}\label{algo:louvain_algorithm}
\end{algorithm}

\section{Multimodal Input Representation}

In our model, each news record $r$ is inputted as a low-dimensional vector $f_{input}(r)$ using its text content (i.e., news title) and propagation network (i.e., social context). Initially, we construct two independent representations for $r$ using its text content $f_{text}(r)$ and propagation network $f_{network}(r)$. Then, these two representations are concatenated to produce the final representation of $r$: $f_{input}(r) = f_{text}(r) \oplus f_{network}(r)$. This process is elaborated in this section.  
\begin{figure}[t]
    \centering
    \includegraphics[width=\linewidth]{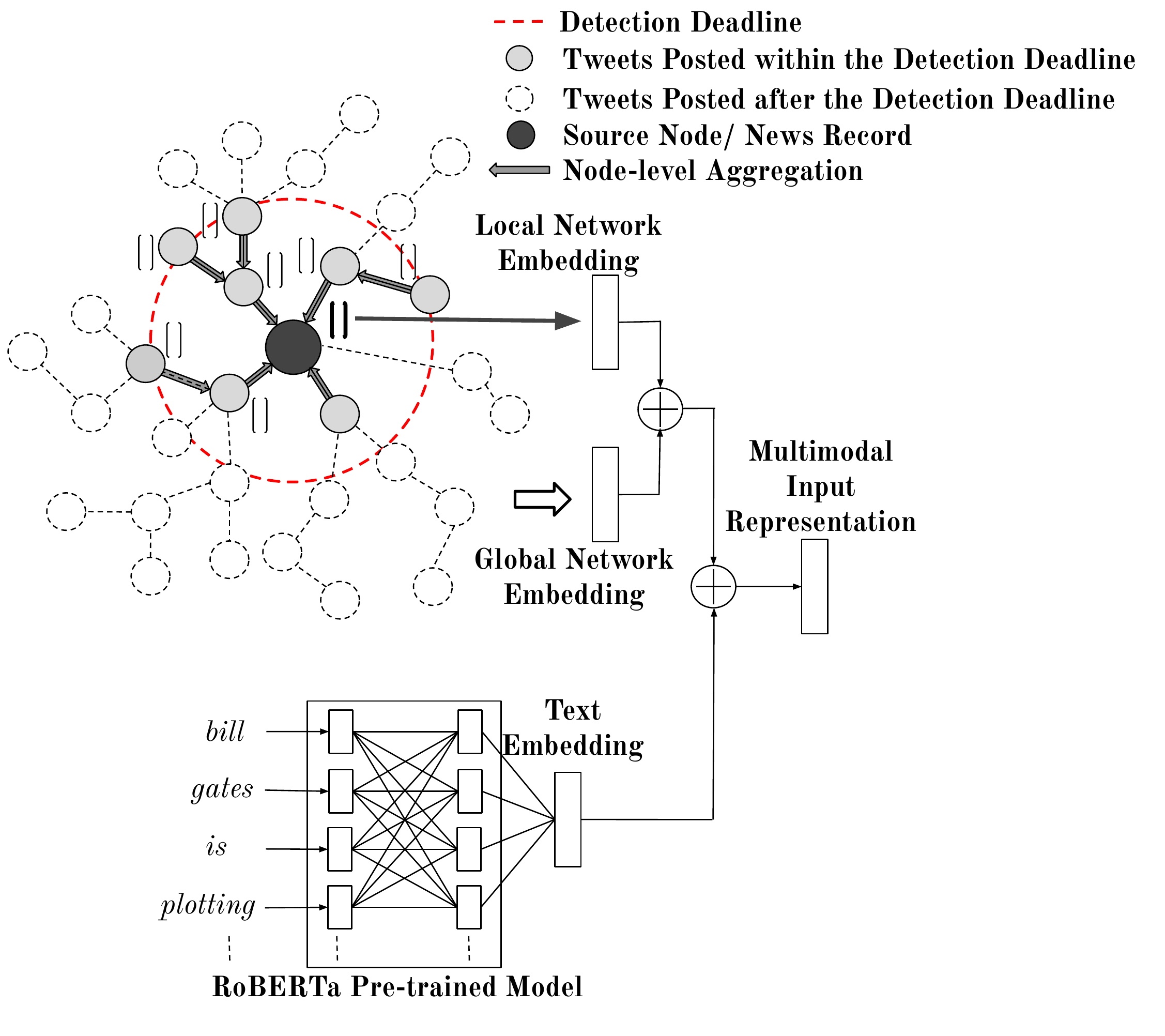}
    \caption{Multimodal Input Representation}
    \label{fig:input_representation}
\end{figure}

\subsection{Text Representation}
In this work, the text content of a news record is represented using RoBERTa~\cite{liu2019roberta}, a robustly optimized BERT pre-training  model. For a given textual content $\{w_1 w_2 w_3 .... w_n\}$ of a news record $r$, the RoBERTa model returns the text-based latent representation $f_{text}(r) \in \mathbb{R}^{d_t}$ of $r$. Out of the different variants of pretrained RoBERTa models, we adopt the roberta-large model available in \url{https://pytorch.org/hub/pytorch_fairseq_roberta/}, where $d_t=1024$.


\subsection{Propagation Network Representation}
We explore two types of features: global-level features (global); and node-level features (local), of the propagation network $G^r=(V^r,E^r,X^r)$ to generate the network-based representation $f_{network}(r)$ of a record $r$. 

\subsubsection{Propagation Network Construction}
We consider all the tweets/retweets related to $r$ as the nodes $V^r$ of $G^r$. There is an extra node (i.e., source node) in $G^r$ to represent the news, which links different information cascades of $r$. The edges $E^r$ of $G^r$ represent how a news item spreads from one person to another as shown in Fig~\ref{fig:input_representation}. Specifically, there is an edge from node \(i\) to node \(j\) if (1) the user of tweet \(i\) mentions the user of tweet \(j\); or (2) tweet \(i\) is public and tweet \(j\) is posted within the detection deadline (= five hours) after tweet \(i\).

\subsubsection{Global Representation} We use the following features as global-level features: (1) Wiener Index ($g_1$); (2) Number of nodes ($g_2$); (3) Network depth ($g_3$); (4) Number of nodes at different hops ($g_5$); and (5) Branching factor at different levels ($g_6$). Finally, all these features are concatenated together to formulate the global-level network representation $f_{global}(r)$ of a record $r$.
 
\begin{table}[t]
    \centering
    \caption{Node-level Features}
    \begin{tabular}{|c|p{6.5cm}|}
         \hline
         Type & Features\\
         \hline
         user & whether the user is verified ($n_1$), the number of followers ($n_2$), the number of friends ($n_3$), the number of lists ($n_4$), and the number of favourites ($n_5$)\\
         \hline
         text & the sentiment scores computed using VADER with the text content in the tweet ($n_6$), the proportion of positive words ($n_7$), the proportion of negative words ($n_8$), the number of mentions ($n_{9}$), and the number of hashtags($n_{10}$)\\
         \hline
         temporal & the time difference with the source node ($n_{11}$); the time difference with the immediate predecessor ($n_{12}$); and the average time difference with the immediate successors ($n_{13}$); user account timestamp ($n_{14}$)\\
         \hline
    \end{tabular}
    \label{tab:node-level_feas}
\end{table}

\subsubsection{Local Representation} For the node-level features, we extract three types of features: (1) text-based; (2) user-based; and (3) temporal-based, which are listed in Table~\ref{tab:node-level_feas}. For a given propagation network $G^r$ of a record $r$, all the features in Table~\ref{tab:node-level_feas} are extracted to represent each vertex (i.e., tweet) in $G^r$. Then, we adopt the node-level aggregation approach proposed in~\cite{silva-embedding-2020} to propagate the aforementioned node-level features to the source node as elaborated in Algo.~\ref{algo:propagation}. This algorithm returns the final representation of the source node (see Fig.~\ref{fig:input_representation}) of $G^r$ as the local representation $f_{local}(r)$ of $r$.

Finally, the network-based representation is formulated as:
\begin{equation}
    f_{network}(r) = f_{global}(r) \oplus f_{local}(r)
\end{equation}
where $\oplus$ denotes concatenation.

\begin{algorithm}[t]
 \LinesNumbered
 \SetAlgoLined
 \SetKwInput{Input}{Input}
 \SetKwInput{Output}{Output}
 \Input{propagation network $G^r=(V^r,E^r,X^r)$\linebreak source node of $r$ $v_s \in V^r$}
 \Output{The local representation $f_{local}(r)$}
$h^0_v \leftarrow x_v \;\; \forall $v$ \in V^r$ \\
 \For{$t$ in ${1,2,...,k}$}{
    \For{$v$ in $V$}{
    $h^t_v \leftarrow \frac{1}{2} h^{t-1}_v + \frac{1}{2}\frac{\sum_{\forall (v,u) \in E^r_t} h^{t-1}_u}{\sum_{\forall (v,u) \in E^r_t} 1}$
    }
 }
 $f_{local}(r) \leftarrow h^k_{v_s}$\\
 $return\text{ }f_{local}(r)$ 
\caption{Local Network Representation}
\label{algo:propagation}
\end{algorithm}

Note: We standardise\footnote{\url{https://scikit-learn.org/stable/modules/generated/sklearn.preprocessing.StandardScaler.html}} each dimension of $f_{network}(r)$ before inputting to the model to stabilise the learning process of our model.

\section{Encoding and Decoding Functions}
In our fake news detection classifier, we have six encoding and decoding functions, $(f_{specific}$, $f_{shared}$, $g_{specific}$, $g_{shared}$, $g_{pred}$, $g_{recon})$. In this work, all these functions are modelled as 2-layer feed-forward networks with sigmoid activation. Formally, we can define a encoding/decoding function $f$ that maps an input $x \in \mathbb{R}^{d_{input}}$ to an output $z\in \mathbb{R}^{d_{output}}$ as:
\begin{equation*}
    z = \sigma(A_2 (\sigma(A_1x+b_1))  + b_2) 
\end{equation*}
where $A_1 \in \mathbb{R}^{(d_{hidden}, d_{input})}$, $A_1 \in \mathbb{R}^{(d_{output}, d_{hidden})}$, $b_1 \in \mathbb{R}^{d_{hidden}}$, and $b_2 \in \mathbb{R}^{d_{output}}$ are trainable parameters. $\sigma$ denotes sigmoid activation. We set $d_{hidden}$ as $\max(d_{input}, d_{output})/2$. For example, assume that $f$ takes inputs of 1024 dimensions and outputs of 128 dimensions. Then, the size of the hidden layer is 512. We leave the optimal neural architecture search for each encoding and decoding function in our model as future work.

\section{Domain Discovery Baseline}
We compare our domain discovery approach with the baseline proposed in~\cite{chen2020proactive}, which assigns hard domain labels for news records based on the users engaged with each news record. For the visualization purpose, we convert these hard domain labels (i.e., one-hot vector) to domain embeddings as they preserve pairwise domain similarity between records~\cite{shu_beyond_2019}. We elaborate the steps that we followed to generate the domain embeddings using this baseline as follows:

\begin{enumerate}
    \item Initially, we construct a network by considering each news record as a node.
    \item Each news record $r$ (i.e., node) is represented using the list of the users $U^r$ tweeting the the particular news record.
    \item The pairwise similarity of nodes is computed for a given two nodes $r_1$ and $r_2$ as:
    \begin{equation*}
        similarity(r_1, r_2) = \frac{|U^{r_1}\cap U^{r_2}|}{|U^{r_1}\cup U^{r_2}|}
    \end{equation*}
    Then $r_1$ and $r_2$ are connected in the graph if $similarity(r_1, r_2)>\alpha$. $\alpha$ is set to $0.4$ following the original paper~\cite{chen2020proactive}. 
    \item The Louvain algorithms is used to identify the communities $C={c_1, c_2, ...}$ in the constructed graph, which yields hard cluster (considered as domains) assignment for each node.
    \item Then each node $r$ can be represented as an one-hot vector $\mathbb{I}^r\in\mathbb{R}^{|C|}$, in which $\mathbb{I}^r_i := \{1\text{ }if\text{ }r \in c_i; 0\text{ } otherwise\}$
    \item Finally, we construct the domain embedding $f_{domain}(r) \in \mathbb{R}^{|R|}$ of $r$ by concatenating the cosine similarity scores of $\mathbb{I}^r$ with other news records:
    \begin{equation*}
        f_{domain}(r) = (\mathbb{I}^r\cdot\mathbb{I}^{r_{0}}) \oplus (\mathbb{I}^r\cdot\mathbb{I}^{r_{1}})...\oplus (\mathbb{I}^r\cdot\mathbb{I}^{r_{|R|-1}})
    \end{equation*}
    where $\oplus$ denotes concatenation operation.
\end{enumerate}
Since this approach considers news records as the nodes of the constructed graph, it is difficult to extend such an approach to learn domain embeddings for new records. In contrast, the proposed approach in this paper constructs its knowledge network using words and users as nodes. Thus, we can generate the domain embeddings for a new record using the words and users related to the new record. Also, our approach considers both text and user information of news records to identify their domain labels. 

\section{Fake News Detection Baselines}
We compare our fake news detection model with seven widely used baselines and their variants:

\newcommand*{\equal}{=}
\newcommand*{\comma}{,}
\definecolor{darkbrown}{rgb}{0.4, 0.26, 0.13}
\begin{figure*}[t]
\centering
 \begin{tikzpicture}
    \begin{groupplot}[group style = {group size = 3 by 1, horizontal sep = 40pt}, width = 4.5cm, height = 4.1cm]
        \nextgroupplot[ title = $\lambda_2\equal10\comma\text{ }\lambda_3\equal5$,
            legend style = {column sep = 0.5pt, legend columns = 3, legend to name = grouplegend,}, ytick={0.0, 0.2, 0.4, 0.6,0.7, 0.7, 0.8, 0.9, 1.0},xtick= {-2, -1, 0, 1, 2, 3, 4}, xlabel=$log_2(\lambda_1)$,ymax=0.92, ymin=0.75, ylabel=F1-score]
\addplot[
    color=blue,
    mark=*,
    ]
    coordinates {
    (-2,0.800)(-1,0.821)(0,0.835)(1,0.834)(2.2,0.827)(3.2,0.835)(4,0.797)
    };
\addplot[
    color=red,
    mark=triangle*,
    ]
    coordinates {
    (-2,0.830)(-1,0.831)(0,0.836)(1,0.839)(2.2,0.842)(3.2,0.835)(4,0.822)
    };
\addplot[
    color=darkbrown,
    mark=square*,
    ]
    coordinates {
    (-2,0.848)(-1,0.855)(0,0.869)(1,0.863)(2.2,0.867)(3.2,0.861)(4,0.857)
    };
    \legend{Politifact, GossipCop, CoAID}
        \nextgroupplot[
        title= $\lambda_1\equal1\comma\text{ }\lambda_3\equal5$,,
        ytick={0.0, 0.2, 0.4, 0.6, 0.7, 0.8, 0.9, 1.0},
        xtick = {-2, -1, 0, 1, 2, 3, 4},
        ymax=0.92, ymin=0.75,
        xlabel=$log_2(\lambda_2)$, ylabel=F1-score]
\addplot[
    color=blue,
    mark=*,
    ]
    coordinates {
    (-2,0.791)(-1,0.797)(0,0.812)(1,0.819)(2.2,0.827)(3.2,0.835)(4,0.831)
    };
\addplot[
    color=red,
    mark=triangle*,
    ]
    coordinates {
    (-2,0.830)(-1,0.830)(0,0.831)(1,0.834)(2.2,0.832)(3.2,0.836)(4,0.831)
    };
\addplot[
    color=darkbrown,
    mark=square*,
    ]
    coordinates {
    (-2,0.850)(-1,0.853)(0,0.855)(1,0.864)(2.2,0.868)(3.2,0.869)(4,0.868)
    };
            \nextgroupplot[
        title= $\lambda_1\equal1\comma\text{ }\lambda_2\equal10$,,
        ytick={0.0, 0.2, 0.4, 0.6, 0.7, 0.8, 0.9, 1.0},
        xtick = {-2, -1, 0, 1, 2, 3, 4},
        ymax=0.92, ymin=0.75,
        xlabel=$log_2(\lambda_3)$, ylabel=F1-score]
\addplot[
    color=blue,
    mark=*,
    ]
    coordinates {
    (-2,0.820)(-1,0.823)(0,0.828)(1,0.835)(2.2,0.835)(3.2,0.818)(4,0.794)
    };
\addplot[
    color=red,
    mark=triangle*,
    ]
    coordinates {
    (-2,0.832)(-1,0.828)(0,0.832)(1,0.832)(2.2,0.836)(3.2,0.839)(4,0.842)
    };
\addplot[
    color=darkbrown,
    mark=square*,
    ]
    coordinates {
    (-2,0.852)(-1,0.857)(0,0.863)(1,0.865)(2.2,0.868)(3.2,0.857)(4,0.846)
    };
    \end{groupplot}
    \node at ($(group c2r1) + (0.0cm,-2.6cm)$) {\ref{grouplegend}};
\end{tikzpicture}
\caption{F1-scores for the fake news detection task with different hyper-parameters: $\lambda_1$; $\lambda_2$; and $\lambda_3$.}
\label{tab:lambda_para}
\end{figure*}
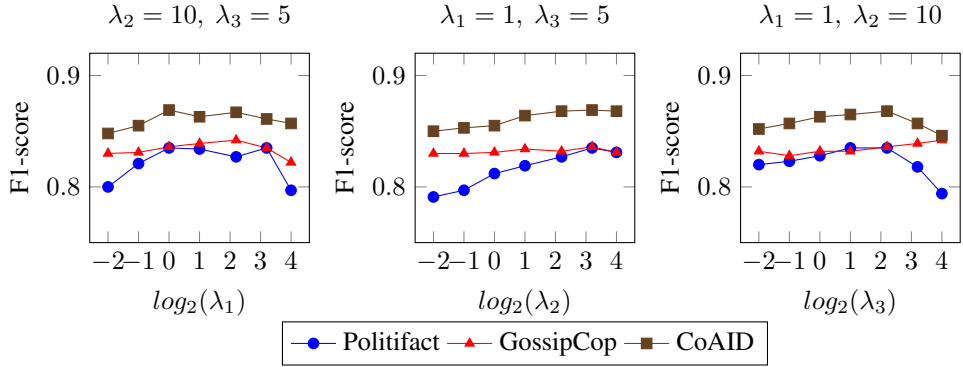

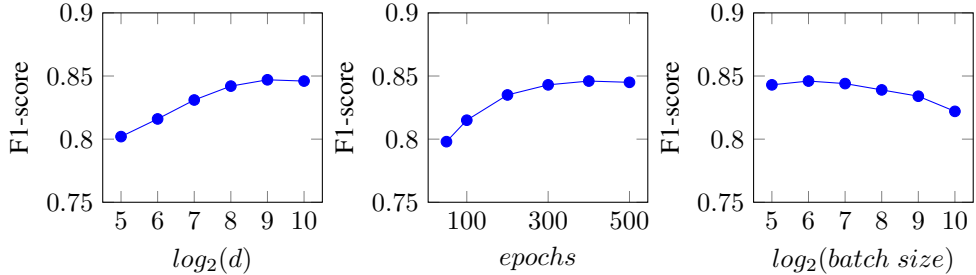
\begin{figure*}[t]
\centering
 \begin{tikzpicture}
    \begin{groupplot}[group style = {group size = 3 by 1, horizontal sep = 40pt}, width = 4.5cm, height = 4.1cm]
        \nextgroupplot[
            legend style = {column sep = 0.5pt, legend columns = 3, legend to name = grouplegend,}, ytick={0.0, 0.2, 0.4, 0.6,0.7, 0.75, 0.8, 0.85, 0.9},xtick= {5, 6, 7, 8, 9, 10}, xlabel=$log_2(d)$,ymax=0.9, ymin=0.75, ylabel=F1-score]
\addplot[
    color=blue,
    mark=*,
    ]
    coordinates {
    (5,0.802)(6,0.816)(7,0.831)(8,0.842)(9,0.847)(10,0.846)
    };
    \legend{Politifact, GossipCop, CoAID}
        \nextgroupplot[
        ytick={0.7, 0.75, 0.8, 0.85, 0.9},
        xtick = {100, 300, 500},
        ymax=0.9, ymin=0.75,
        xlabel=$epochs$, ylabel=F1-score]
\addplot[
    color=blue,
    mark=*,
    ]
    coordinates {
    (50, 0.798)(100,0.815)(200,0.835)(300,0.843)(400,0.846)(500,0.845)
    };
            \nextgroupplot[
        ytick={0.7, 0.75, 0.8, 0.85, 0.9},
        xtick = {5, 6, 7, 8, 9, 10},
        ymax=0.9, ymin=0.75,
        xlabel=$log_2(batch\;size)$,
        ylabel=F1-score]
\addplot[
    color=blue,
    mark=*,
    ]
    coordinates {
    (5,0.843)(6,0.846)(7,0.844)(8,0.839)(9,0.834)(10,0.822)
    };
    \end{groupplot}
\end{tikzpicture}
\caption{F1-scores (overall for the all three datasets) for the fake news detection task with different hyper-parameters: $d$; $epochs$; and $batch\text{ }size$.}
\label{tab:different_para}
\vspace{-5mm}
\end{figure*}

\begin{itemize}
    \item LIWC~\cite{pennebaker_development_2015} ((i.e., Linguistic Inquiry and Word Count)) learns feature vectors from the text content of news records by counting the number of lexicons falling into different psycho-linguistic categories\footnote{\url{https://liwc.wpengine.com/}}. Then, a logistic regression model\footnote{\url{https://scikit-learn.org/stable/modules/generated/sklearn.linear_model.LogisticRegression.html}} is used as the classifier to predict fake news using LIWC feature vectors.  
    \item text-CNN~\cite{kim_convolutional_2014} uses Convolution Neural Networks (CNN) to model  the text content of news records at different granularity levels with the help of multiple convolutional filters and multiple CNN layers\footnote{\url{https://github.com/yoonkim/CNN_sentence}}.
    \item HAN~\cite{yang_hierarchical_2016} adopts a hierarchical attention neural network framework to model the text content of news records, which can assign varying importance to words and sentences when making final predictions by word-level and sentence-level attention\footnote{\url{https://github.com/tqtg/hierarchical-attention-networks}}.
    \item EANN~\cite{wang_eann_2018} produces a latent representation for each news record using its different modalities (e.g., text, network) such that the domain-specific knowledge in news records are ignored in the latent space. Subsequently, the latent representation is used to predict the label of the news record. We compare our model with two variants of EANN:
    \begin{itemize}
        \item EANN-Unimodal only considers the text modality of a news record to generate the latent representation; and
        \item EANN-Multimodal considers both text and network modalities of a news record to produce the latent embedding.
    \end{itemize}
    For a fair comparison of the models, we adopt the same text and network representation techniques in our model to encode the input modalities of EANN.
    \item HPNF~\cite{shu_hierarchical_2019} extracts various features (e.g., structural features, temporal features) from the propagation network of a news record to generate its feature representation. Then, a Logistic Regression is used to classify news records using the extracted propagation network-based model. In HPNF+LIWC, we concatenate the features vectors from HPNF and LIWC together to construct the feature representation for news records.     
    \item AE~\cite{silva-embedding-2020} adopts an Auto-encoder architecture to learn latent representation for each news record based on its propagation network. Subsequently, the latent representations are used to determine fake news records.
    \item SAFE~\cite{zhou_safe_2020} proposes a multimodal approach for fake news detection. For a given news record, this model learns separate latent representations for each modality. Also, it jointly learns another representation to represent cross-modality knowledge, which is consistent across modalities. Finally, all three representations are concatenated and fed to a classifier to predict the label of the record. The original work of this model considers the text and image modalities of news records. For a fair comparison with our model, here we use the text and network modality of news records for this baseline too. We adopt the same text and network representation techniques in our model to encode the input modalities in this baseline too.
\end{itemize}

\section{Parameter Sensitivity}
This section evaluates how changes to the hyper-parameters of the model affect its performance on the fake news detection tasks.

In Figure~\ref{tab:lambda_para}, we analyse the performance of our model for different $\lambda_1$, $\lambda_2$ and $\lambda_3$ values (see Eq. 7 in the paper), which varies the importance assign to each loss term in our model. By setting a very high value ($>2^2$)
or a very low value ($<2^{-1}$) for $\lambda_1$ tends to drop the performance consistently for all three datasets. It means that $L_{recon}$ loss term should be included in our model with moderate importance compared to the other loss terms. The performance of the model for PolitiFact and CoAID domains drop substantially for low $\lambda_2<5$ and high $\lambda_3>5$ values. By setting a low $\lambda_2<5$ or a high $\lambda_3>5$ value, our model assigns more importance to the cross-domain embedding space. The cross-domain embedding space could be dominated by frequently appearing domains (GossipCop in this dataset). Thus, assigning more importance for cross-domain embedding space, the model could poorly perform for small domains e.g., PolitiFact and CoAID in this dataset as shown in Fig.~\ref{tab:lambda_para}. This observation further signifies the importance of having domain-specific knowledge of news items to identify fake news.

We examine the sensitivity of the model's performance for other parameters: latent dimension $(d)$; number of epochs; and batch size. Overall, the model yields consistent performance for $d>256$, $epochs>300$, and $batch\text{ }size<128$ values.   

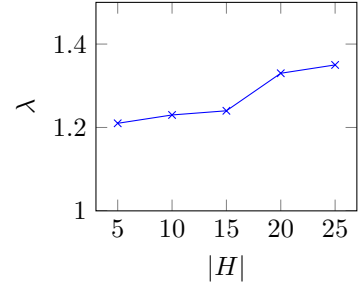
\begin{figure}[t]
\centering
 \begin{tikzpicture}

    \begin{axis}[
		xlabel=$|H|$,
		ylabel=$\lambda$,
		width=0.6\linewidth,
		xtick={5, 10, 15, 20, 25},
		legend style={at={(0.5,-0.40)},
    anchor=north,legend columns=2}, ymax=1.5, ymin=1]
	\addplot[color=blue,mark=x] coordinates {
		(5,1.21)
		(10,1.23)
		(15,1.24)
		(20,1.33)
		(25,1.35)
	};
	\end{axis}
\end{tikzpicture}
\caption{Domain-coverage measure $\lambda$ (lower $\lambda$ is better) of the dataset selected using the LSH-based instance selection with different $|H|$ (number of hash functions) values, when $B/|R|=0.1$.}
\label{fig:different_h_values}
\end{figure}

There is only one hyper-parameter in the proposed LSH-based instance selection approach, which is the number of hash functions ($|H|$) used for the random projections. As shown in Figure~\ref{fig:different_h_values}, domain coverage of the proposed approach reduces (increases $\lambda$ measure) for high $|H|(>20)$ values. This is intuitive because high $|H|$(lengthy hash codes) value could map even very close neighbours in the embedding space into different bins. Thus, the selected instance from different bins could be close-neighbours. In contrast, low $|H|$ values increases the domain coverage. Nevertheless, having a very low $|H|$ value increases the time complexity as it requires many iterations of the hashing step to meet a given labelling budget.

In summary, we adopt the following hyper-parameter values for the results reported in the paper: (1) $lambda_1=1$; (2) $lambda_2=10$; (3) $lambda_3=5$; (4) $|H|=10$; (5) $d=512$; (6) $epochs=300$; (7) $batch\;size=64$. We use the Adam optimizer for the optimization. For the parameters of the optimizer (e.g., learning rate, moments), the default parameters in Keras\footnote{\url{https://keras.io/api/optimizers/adam/}} are used. Due to the randomness involved in the training and testing datasets splitting process, we conducted all our experiments using three random state value: $\{0,1,2\}$, and the average performance is reported in the paper. 

\begin{figure}[t]
\centering
\begin{tikzpicture}
\begin{axis}[
        xlabel=$epochs$,
		ylabel=$loss\;value$,
		width=0.9\linewidth,
		height=0.7\linewidth,
		xmin=5,
		xtick={25, 100, 200, 300, 400, 500},legend cell align={left}
]
\addplot[color=black,mark=none] table [x=a, y=b, col sep=comma] {data3.csv};
\addplot[color=blue, mark=none] table [mark=none, x=a, y=c, col sep=comma, color=blue] {data3.csv};
\addplot[color=red,mark=none] table [mark=none, x=a, y=d, col sep=comma] {data3.csv};
\addplot[dashed, thick, color=blue,mark=none] table [mark=none, x=a, y=e, col sep=comma] {data3.csv};
\addplot[dashed, thick, color=red,mark=none] table [mark=none, x=a, y=f, col sep=comma] {data3.csv};
\legend{$L_{final}$,$L_{pred}$,$\lambda_1L_{recon}$,$\lambda_3L_{specific}$,$\lambda_2L_{shared}$}
\end{axis}
\end{tikzpicture}
\caption{Convergence properties of the loss function.}
\label{fig:convergence_analysis}
\vspace{-5mm}
\end{figure}
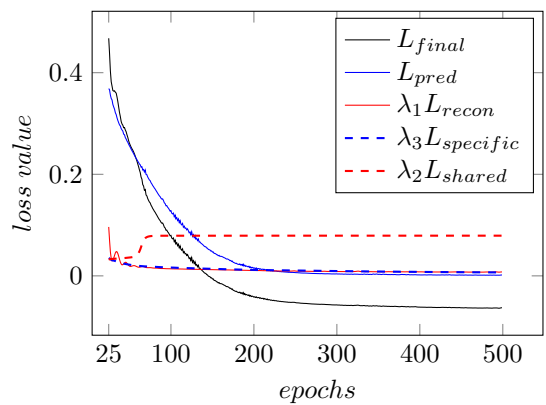

\section{Convergence Analysis}
In Figure~\ref{fig:convergence_analysis}, we examine the convergence properties of the loss function of our model. Our loss function consists of four terms: prediction loss ($L_{pred}$); reconstruction loss ($L_{recon}$); domain-specific loss ($L_{specific}$); and cross-domain loss ($L_{shared}$). As can be seen in Fig.~\ref{fig:convergence_analysis}, each loss term converges around 250 epochs. Since $L_{shared}$ is trained as a minimax game, the converging $L_{shared}$ in Fig.~\ref{fig:convergence_analysis} empirically verifies the convergence of the proposed minimax game to exploit cross-domain knowledge in news records. Moreover, $L_{recon}$, $L_{specific}$ and $L_{shared}$ are mean-squared error based loss terms and $L_{pred}$ is based on binary cross-entropy. Hence, the typical value range for the non-converged $L_{pred}$ differs from the other loss terms. This also shows the importance of having $\lambda_1$, $\lambda_2$, and $\lambda_3$ to penalise such differences due to different loss functions. 

\bibliography{Supplement.bib}